 \documentclass[accepted]{uai2022} 

\usepackage[american]{babel}

\usepackage{natbib} 
\usepackage{mathtools} 
\usepackage{booktabs} 
\usepackage{tikz} 
\usepackage{amsthm}
\usepackage{diagbox}
\usepackage{subcaption}
\usepackage{adjustbox}
\usepackage{afterpage}
\usepackage{authblk}
\usepackage{graphicx}
\usepackage{hyperref}       
\usepackage{url}            
\usepackage{booktabs}       
\usepackage{amsfonts}       
\usepackage{nicefrac}       
\usepackage{microtype}      
\usepackage{xcolor}         
\usepackage{multirow}
\usepackage{booktabs}
\usepackage{hhline}
\usepackage{algorithm}
\usepackage{algpseudocode}
\newcommand{\Var}{\mathrm{Var}}
\newcommand{\MSE}{\mathrm{MSE}}
\newcommand{\MMSE}{\mathrm{MMSE}}

\newtheorem{prop}{Proposition}
\newtheorem{conj}{Conjecture}
\newcommand{\indep}{\perp \!\!\! \perp}
\usepackage{hyperref}
\usepackage{authblk}
\definecolor{mydarkblue}{rgb}{0,0.08,0.55}
\hypersetup{  
    colorlinks=true,
    linkcolor=mydarkblue,
    citecolor=mydarkblue,
    filecolor=mydarkblue,
    urlcolor=mydarkblue
}

\title{Tearing Apart NOTEARS: \\Controlling the Graph Prediction via Variance Manipulation}

\author[1]{Jonas Seng}
\author[1]{Matej Zečević}
\author[1,3]{Devendra Singh Dhami}
\author[1,2,3]{Kristian Kersting}
\affil[1]{Computer Science Department and \textsuperscript{\rm{2}}Centre for Cognitive Science, TU Darmstadt}
\affil[3]{Hessian Center for AI (hessian.AI) \quad \textsuperscript{\rm{$\dagger$}}correspondence:\ \texttt{jonas.seng@tu-darmstadt.de}}

\begin{document}
\maketitle
\begin{abstract}
Simulations are ubiquitous in machine learning. Especially in graph learning, simulations of Directed Acyclic Graphs (DAG) are being deployed for evaluating new algorithms. In the literature, it was recently argued that continuous-optimization approaches to structure discovery such as NOTEARS might be exploiting the sortability of the variable's variances in the available data due to their use of least square losses. Specifically, since structure discovery is a key problem in science and beyond, we want to be invariant to the scale being used for measuring our data (e.g. meter versus centimeter should not affect the causal direction inferred by the algorithm). In this work, we further strengthen this initial, negative empirical suggestion by both proving key results in the multivariate case and corroborating with further empirical evidence. In particular, we show that we can control the resulting graph with our targeted variance attacks, even in the case where we can only partially manipulate the variances of the data.
\end{abstract}

\section{Introduction}\label{sec:intro}
Given a finite data sample from an unknown probability distribution, structure learning algorithms aim to recover the graphical structure underlying the data generating process that lead to said unknown probability distribution (for an introduction to probabilistic graphical models see \citep{koller2009probabilistic}). These structure learning (or graph learning or graph induction) algorithms are being deployed widely throughout many Machine Learning applications because of their efficacy in representing the probability distributions compactly and (to some extent) interpretable, for instance, Bayesian Networks (BN; see \citet{pearl2000bayesian}). In many cases, including BNs, a Directed Acyclic Graph (DAG) is being used as representation of choice. In DAGs a node corresponds to a random variable and each edge marks a direct statistical dependence between two random variables. The absence of an edge encodes (in)direct independencies between random variables. A natural extension to BNs are Structural Causal Models (SCM; see \citet{pearl2009causality}) which imply a causal graph i.e., an edge now refers to a causal relationship amongst adjacent variables.

Recovering the independence-structure of a probability distribution based on finite samples is not a trivial task and many different approaches have been proposed to solve this task. Some use statistical independence-tests to infer a graph, others use score-functions which are being optimized during learning (for a more complete overview including the causal perspective consider \citep{mooij2020joint} or \citep{peters2017elements}). One major problem for score-based approaches is to ensure that the resulting graph is a valid DAG i.e., to ensure that there are no cycles in the produced graph. NOTEARS \citep{zheng2018dags} is a recent score-based structure learning algorithm which introduces a continuous and differentiable DAG-constraint opposed to the otherwise combinatorial, constructive constraint. This so-called acyclicity constraint takes on value 0 iff an adjacency matrix (a matrix-representation of a graph) is a DAG. NOTEARS yields state of the art results for many structure learning tasks and even recovering causal structures from observational data seemed to be solved by NOTEARS in cases where the dependencies between variables are linear and the variables follow a Gaussian distribution. This result is surprising since before NOTEARS it was already proven theoretically that in such cases identification of causal structures is impossible \citep{shimizu2006linear}.

\cite{reisach2021beware} argued that NOTEARS's ability to recover causal structures, for which it has been proven impossible given only the observational data, is due to a property they called \textit{varsortability}, at least as long as a least square based loss function is used as an optimization objective. Following their arguments, NOTEARS prefers structures which resemble the causal structure of some data generating process because the \textit{var}iances of Guassian variables are added up if one follows the causal path in a data generating process, thus making the nodes \textit{sortable} based on their variance. They have shown that in such cases the means squared error (MSE) used by NOTEARS is smaller for a model respecting the true causal structure of the data than for a model which does not. In a nutshell, this would imply that choosing your scale affects the causality detected to be underlying the data which obviously is nonsensical. 
From now on we will call the NOTEARS formulation using least square based losses \textit{Standard NOTEARS} (SNT).
\begin{figure}
    \centering
    \includegraphics[width=.5\textwidth]{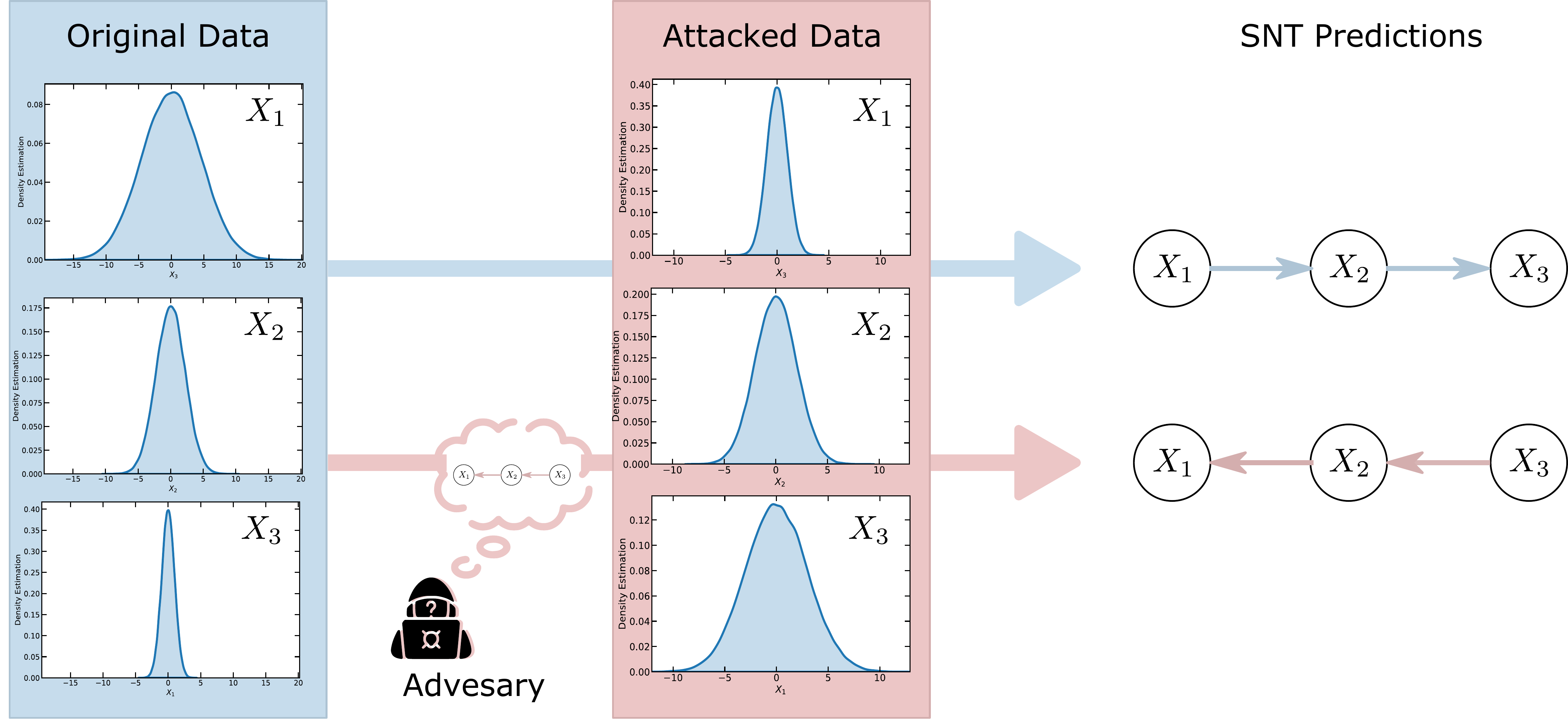}
    \caption{\textbf{Manipulating the Data Variances Through Rescaling.} This illustration shows the basic concept of our attacks. First we sample data from a SCM (left, in this case a chain-structure), thus obtaining data with varsortability 1 w.r.t. to the true causal graph (left density plots). By rescaling, we can change the varsortability-property w.r.t. the true graph arbitrarily, thus changing the varsortability w.r.t. a graph we want SNT to predict as well (graph with red box on the right). Since SNT prefers graphs with high varsortability, we are able to change the structure predicted by SNT by rescaling data (right density plots). \small{(Best viewed in color.)}}
    \label{fig:tant_concept}
\end{figure}
Our contribution goes a step beyond this and shows that it is possible to predict the graph SNT will recover from data by manipulating the variance of the data appropriately in a linear Gaussian setting. We choose SNT specifically as it characterizes the key properties of a current family of continuous optimization discovery algorithms i.e., it only uses three components (i) a least square based loss, (ii) a continuous acyclicity constraint and (iii) a regularizer. Furthermore, it is a method that has gained recognition in the community being deployed widely in application settings (see software packages like \citep{Beaumont_CausalNex_2021} and \citep{zhang2021gcastle} or follow-up works like \citep{yu2019dag,lee2019scaling, wei2020dags}), making it so important that the method is well-understood. We look at the multivariate case of graphs with at least three nodes, as it is of most practical interest. There specifically we are considering the three settings: Manipulating chain-structures, forks and colliders. Thus we will show that any possible node-configuration in a graph can be attacked in certain ways. Additionally we will provide theoretical justifications for our attacks. We make our code publicaly available at: {\footnotesize \url{https://anonymous.4open.science/r/TANT-8B50/}}.


\paragraph{Related Work.} To the best of our knowledge \citep{reisach2021beware} were the first to start raising awareness towards the issue of standardization leading to algorithm performance degeneration (alongside \citep{kaiser2021unsuitability} who independently reported similar results on the failure cases). Therefore, this present work tries to strengthen the previous claims, raising more awareness, thereby seeing itself as a direct follow-up to \citep{reisach2021beware}. Important works that discuss identifiability and its relation to variance (upon which also the previously mentioned build) include \citep{peters2014identifiability, park2020identifiability, weichwald2020causal}. There has also been works on describing the limitations of NOTEARS in causal discovery tasks \citep{he2021daring,gao2021dag}.


\section{Methods \& Theoretical Foundations}\label{sec:methods}
In the following we consider a Structural Causal Model (SCM; see \citep{pearl2009causality,peters2017elements}) $\mathcal{M}$ with an associated causal graph $G$ and a probability distribution $P$ that factorizes according to $G$ (Markov factorization property which is equivalent to the statement that $d$-separation implies conditional independence since $P$ is assumed to have a density). We assume that we are given a dataset $\mathcal{D}$ of $n$ i.i.d. samples from $P$.

We note that there are exactly three structures that any causal graph is composed of: (1) chain structures consist of $n$ single-link structures where one single-link structure is followed by another, i.e. $X_1 \rightarrow X_2 \rightarrow ... \rightarrow X_n$, then (2) fork structures share the same $d$-separation statements as chains do, however their graph structure is slightly different and is defined as $X_1 \leftarrow ... \leftarrow X_{i-1} \leftarrow X_i \rightarrow X_{i+1} \rightarrow ... \rightarrow X_n$ and finally (3) colliders given by $X_1 \rightarrow X_2 \leftarrow X_3$.

We note that our attacks are only able to turn edges in the graph, it is not possible to add new (in)dependencies in the data with these attacks since only the variances are being changed. However, it is still possible that as a consequence of our attack new edges appear in the attacked graph due to the rules of $d$-separation and Markovianity of $P$ w.r.t. $G$ to satisfy the existing independencies in the data.

\subsection{Attack Definition}
For our attack we assume that: (1) The data $\mathcal{D}$ contains samples from each variable in the SCM, (2) each function $f: \mathbf{PA}_{X_i} \rightarrow X_i$ is a linear function with additive noise where $\mathbf{PA}_{X_i}$ are the parents of a variable $X_i$ and (3) we have full control over $\mathcal{D}$. This means we can measure the variance of each variable in $\mathcal{D}$ as well as manipulate the variance of the data of each variable. Formally, we describe the attack as follows: Given a causal graph $G$ and a target graph $G'$, both having the same set of nodes $\{X_1, \dots, X_d\}$, as well as a distribution $P$ over $G$, we select each node $X_i$ in $G'$, obtain its parents $\mathbf{P}_{X_i}^{(G')}$ in $G'$ and scale data using a simple rule:
\begin{align}\label{eq:attack}
\begin{split}
    & \mathbf{x}_i = c \cdot \mathbf{x}_i \\
    \text{s.t. } & \forall X_j \in \mathbf{P}_{X_i}^{(G')}: \Var(X_i) > \Var(X_j)
\end{split}
\end{align}
Here, $\mathbf{x}_i$ refers to data sampled from node $X_i$ in $G$.
An attack is considered to be successful if SNT predicts a graph $\hat{G}$ s.t. $G' = \hat{G}$ and $G' \neq G$.

Since NOTEARS in general is independent of the choice of the loss function, note that our attacks might fail once we replace the loss by an objective which is not least squared based. We do argue that least squares (as long as we work in the space of DAGs) might be the defining component to susceptibility to variance manipulations, therefore, we might expect similar behavior not just from methods that build upon SNT (for instance \citet{yu2019dag,wei2020dags}; as indicative results in \citep{reisach2021beware} seemed to suggest) but that simply share these two properties of least squares and DAG hypothesis space. A rigorous investigation of such, more general settings is left for future work.


We coin this type of manipulation ``attack'' since it involves a \emph{targeted} manipulation of the system i.e., a priori a target graph is formulated which should be predicted on the data that has been attacked.

\subsection{Theoretical Foundation}
To our conviction, the main argument in the discussion around SNT is that it only minimizes MSE under a DAG hypothesis-space, thereby also lacking capabilities of inferring the underlying causal structure. Additionally in this work we will give rise to assume that minimizing the MSE is unique and equivalent to varsortability being one. We will apply a trick to support this conjecture:
Remember, varsortability is computed from the graphical model modelling the distribution the actual data is sampled from:
Given a causal graph $G$ over variables $\{X_1, \dots, X_d\}$ and an adjacency matrix $\mathbf{A}$ representing $G$, varsortability is defined as the fraction of directed paths that start from a node with strictly lower variance than the node they end in:
\begin{equation}
    v := \frac{\sum_{k=1}^{d-1} \sum_{i \rightarrow j \in \mathbf{A}^k} \text{inc}(\Var(X_i), \Var(X_j))}{\sum_{k=1}^{d-1}  \sum_{i \rightarrow j \in \mathbf{A}^k} 1}
\end{equation}
Here, inc is defined as:
\begin{equation}
    \text{inc}(a, b) := \begin{cases}
1 & a < b \\
\frac{1}{2} & a = b \\
0 & a > b
\end{cases}
\end{equation}
The above definition was taken from \cite{reisach2021beware}.
In this case, a varsortability of one implies that varsortation is equivalent to the correct causal order of variables. We will show that we can exploit this property by ``redefining'' the causal order. For example, assume a causal graph $X_1 \rightarrow X_2 \rightarrow X_3$. If we obtain data $\mathcal{D}$ with varsortability of one, this implies that the varsortation corresponds to the correct causal order. This also means that the varsortability of $\mathcal{D}$ w.r.t. to $X_1 \leftarrow X_2 \leftarrow X_3$ equals to zero. However, if we scale data s.t. varsortability equals one for the last graph, SNT will prefer $X_1 \leftarrow X_2 \leftarrow X_3$ as the correct solution.

Returning to our attack definition Eq. \ref{eq:attack}, we note that \citet{reisach2021beware} looked at a special case of manipulation, namely standardization. However, we do not classify this manipulation as ``attack'' since they were not concerned with actually changing the result to a certain \emph{target}.
We rephrase the key result from \cite{reisach2021beware} formally for the bivariate case:
\begin{prop}\label{prop:ReisachProof}
Given a causal graph $X_1 \rightarrow X_2$ and data from a distribution $P_{X_1 \rightarrow X_2}$, SNT finds the correct graph iff. $\Var(X_2) > \Var(X_1)$. \qed
\end{prop}
Our first theoretical insight involves extending the chain scenario to the multivariate case:
\begin{conj}\label{conj:conj1}
Consider a $n$-dimensional chain graph, then it suffices to change the variance of the root node to be greater than the sink node, $\Var(X_1)>\Var(X_n)$, to prefer the reverse chain in terms of MSE.
\end{conj}
This conjecture is based on the fact that unrolling the inequality will reveal terms $\Var(X_{i+1}-v_{i}X_i)$ that cancel each other out. However, note that the strong statement of the reverse chain being preferred over all the other possible DAGs is not made. I.e., only the reverse chain is preferred over the ground truth chain, but still a flip on e.g. the last edge will still be better than the reverse chain MSE-wise. This we corroborate empirically, both for this simple case directly but also indirectly with results of transforming a chain to a collider. We will give empirical evidence for this conjecture in Section \ref{sec:Results}. \\
We proceed with another theoretical foundation for our attacks:

\begin{prop}\label{prop:Chain2Collider}
Given a causal graph $X_1 \rightarrow X_2 \rightarrow X_3$ and data $\langle \mathbf{x}_1, \mathbf{x}_2, \mathbf{x}_3 \rangle$ from a distribution $P_{X_1 \rightarrow X_2 \rightarrow X_3}$, SNT predicts a graph in which $X_2$ is a collider and which contains an additional edge between $X_1$ and $X_3$ if we scale $\mathbf{x}_2$ s.t. $\Var(\mathbf{x}_1) < \Var(\mathbf{x}_2)$ and $\Var(\mathbf{x}_3) < \Var(\mathbf{x}_2)$.
\end{prop}
A proof is given in Appendix \ref{proof:prop2}. \qed \\

\section{Perfect Attacks vs Imperfect Attacks}
After establishing theoretical foundations, we will consider two concrete attack scenarios.

\subsection{Perfect Attacks}\label{subsec:perfect}
In the following we will refer to \textit{Perfect Attacks} to scenarios in which we have data $\mathcal{D}$ from some SCM with corresponding graph $G$ and we have full control over $\mathcal{D}$, that is we can measure and alter variances as we need it. In order to make notation easier, we will refer to a dataset obtained from some SCM as $\mathbf{X}$ and we will denote the data-vector of some variable $X_i$ in the SCM as $\mathbf{x}_i$.
As mentioned in Section \ref{sec:methods}, there are three substructures each DAG, and thus each causal graph, consists of: Chains, forks and colliders. All of these structures can be attacked in at least two ways. Chains allow for three attacks: Reversing a chain, introducing a collider and making a chain a fork. All attacks can be employed by simply scaling the variables s.t. the variables can be sorted w.r.t. their variance values. Forks can be attacked by making them a chain or introducing a collider. Changing the forks origin node, i.e. attacking a fork-structure s.t. a different fork structure than the ground truth is predicted by SNT, is not possible since we then had to attack the dependencies in the data itself. \\
Attacking colliders can be done by making them a chain or a fork. As for forks, we cannot change the colliders' sink node due to our restriction of manipulating the variance of the data only, not the dependencies itself. However, by attacking a collider, we achieve a weaker variant of changing the colliders' sink node as a side-product: Since in a collider there is a conditional dependence between two nodes, forcing SNT not to predict the true collider structure, we "observe" the collider's sink node, thus forcing SNT to add an additional dependence. Thus, in the 3-node case, we get three edges in our graph predicted by SNT, leading to a collider in any case. Further details on the attacks and their influence on SNT-predictions can be found in the supplementary material.

\subsection{Imperfect Attacks}\label{subsec:imperfect}
In real world attack-scenarios there might be cases in which a dataset is not fully observable or fully accessible to the attacker. For example, data could be stored in a distributed fashion s.t. each storage-party holds a certain set of features only. Then, the attacker might only have access to a subset of the features, thus unable to control the entire dataset. We term such attacks \textit{Imperfect Attacks}. We now investigate if we still can use our theoretical results and methods to attack SNT, at least s.t. we partially can reach our desired goal. \\
In the imperfect scenario we consider a dataset $\mathbf{X} = \langle \mathbf{x}_1, \dots, \mathbf{x}_m \rangle$ where each $\mathbf{x}_i$ is a vector of $n$ instantiations of some random variable $X_i$ which is part of a SCM. The attacker has access to a subset of features, i.e. $\mathcal{A} \subset \{1, \dots, m\}$. We denote the data the attacker can access and thus manipulate by $\mathbf{X}(\mathcal{A})$.
We will consider the 3-node case only in this work in order to show that some attacks are still possible under these conditions. \\
Note that it is sufficient to have control over $n - 1$ variables in the data sampled from an SCM with $n$ variables in order to render the perfect attack scenario, thus such cases will not be part of consideration here. \\
As for perfect attacks, we can attack chains, forks and colliders in the imperfect scenario. For chains it is guaranteed that we can introduce a collider or a fork as long as we have given access to the data of at least one endogenous variable since the attack would be the same as in the perfect scenario. Reversing a chain is also possible in some cases if we are allowed to manipulate the variance information of one of the exogenous variables. In such cases the success of an attack depends on the scale we chose for the attack, the regularization parameter $\lambda$ chosen for SNT and the strength of dependence between variables. We will give empirical evidence to this in the next section. \\
Forks can be attacked by introducing a collider if we have given access to the fork's origin node and by turning a fork into a chain, which requires access to one of the exogenous nodes. Similarily colliders can be attacked by transforming the collider into a fork if we have access to the collider's sink node or we can introduce a chain if we have access to one of the exogenous nodes. Further details are shown in the supplementary material.
\begin{figure}
    \centering
    \includegraphics[scale=0.4]{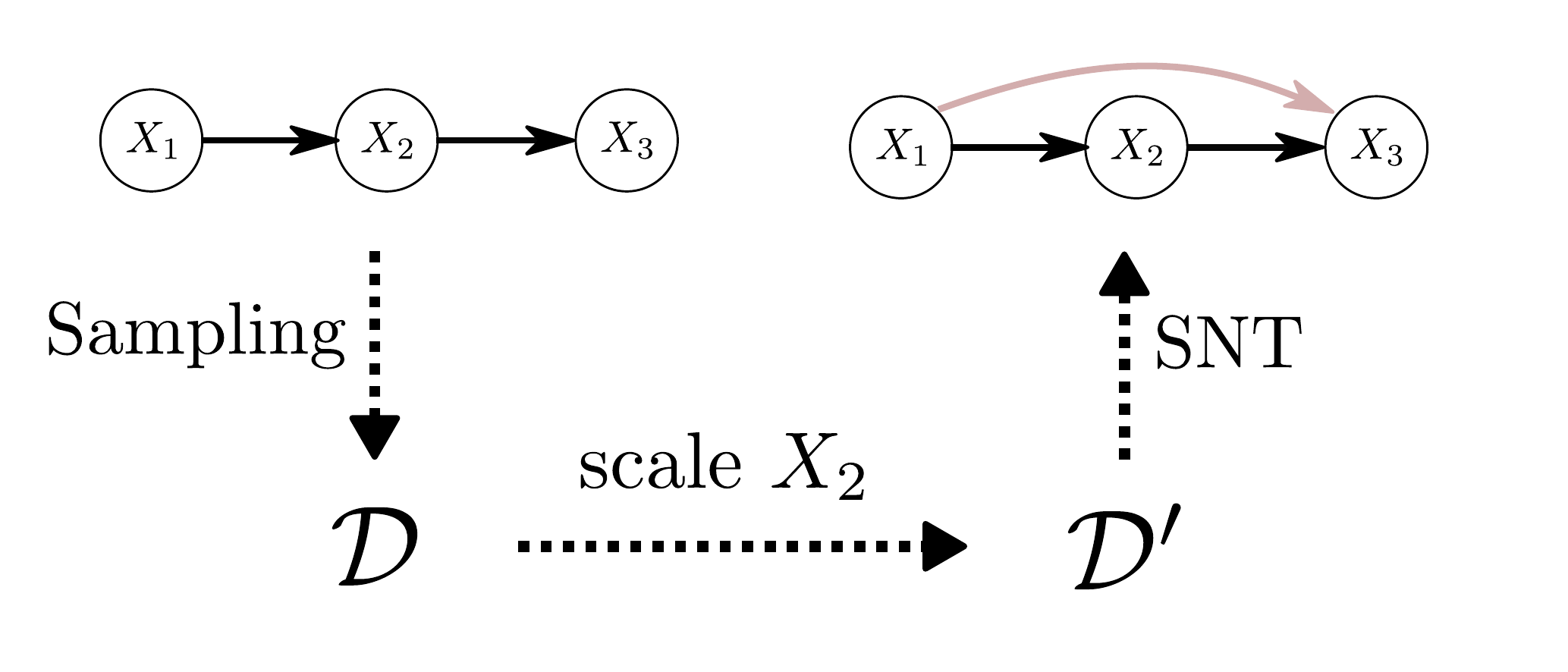}
    \caption{\textbf{Converting Chains to Colliders.} This illustrates the attack on chain-structures which introduce a collider. The green node is the attacked node, i.e. $X_2$ in this case. First data is sampled from the SCM of the corresponding causal graph (left), then we scale $X_2$ in the data and apply SNT. SNT will add an additional edge $X_1 \rightarrow X_3$ (red).}
    \label{fig:attacking_chains_perfect}
\end{figure}

\section{Empirical Results} \label{sec:Results}

In this section the we will show successful attacks on simulated data in both scenarios described in the last section. With this we aim to justify our theoretical findings with empirical data.

\subsection{Generating Data}\label{subsec:data_generation}
In order to back our propositions and conjecture, we employed experiments. Therefore we defined a causal graph $G$ and sampled $10000$ samples from a Gaussian distribution with $\mu = 0$ and a standard deviation $\sigma$ for each exogenous variable $X_i$. Additionally we defined a linear function with additive Gaussian zero-centered noise for each endogenous variable $X_j$ by fixing a weight $w_{i \rightarrow j}$ for each parent of $X_j$, thus computing:
\begin{equation}
    X_j = \sum_{X_i \in \mathbf{PA}_{X_j}} w_{i \rightarrow j} \cdot X_i
\end{equation}
This way we obtain a dataset $\mathbf{X}$. We use $\mathbf{X}$ to apply SNT without manipulating the data and to apply SNT on attacked data. Then we compared the predicted graph in order to see if the attack leads to different output of SNT and if the attack was successful. 

\subsection{Perfect Attacks}
We performed all attacks on chains, forks and colliders as described in Section \ref{subsec:perfect}. We were able to confirm our theoretical findings and obtained a success-ratio of $1$ for all attacks on chsins, forks and colliders. Thus we have also shown empirically that the output of SNT is fully predictable in terms of our attacks.
\begin{figure}[t]
    \centering
    \includegraphics[scale=0.3]{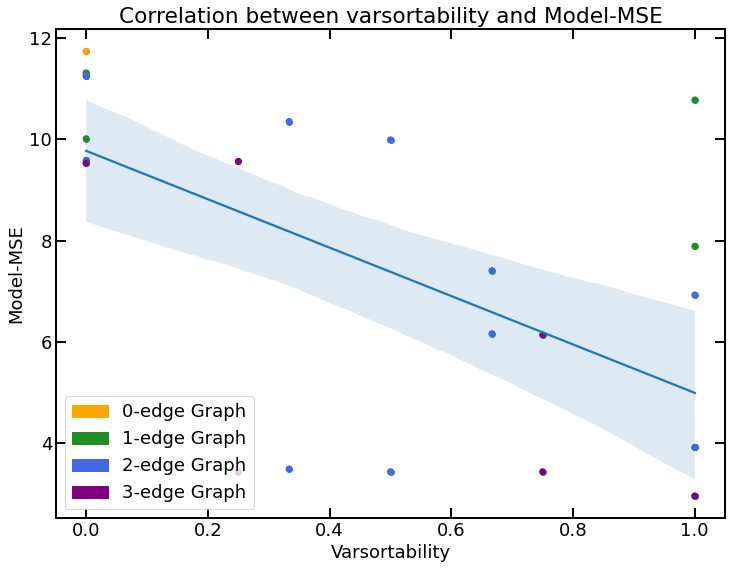}
    \caption{\textbf{Correlation between varsortability and MSE.} We fit a linear model according to each possible DAG in the 3-node case and compute the MSE and varsortability for each. The underlying data-generating system was a chain. It shows a correlation (blue line with 95\%-confidence interval) between varsortability and the Model-MSE. Additionally we see that more complex graphs have lower MSE and higher varsoratbility. \small{(Best viewed in color.)}}
    \label{fig:var_mse_correlation}
\end{figure}
\subsection{Imperfect Attacks}
Starting with attacks on chains, we observed that reversing a chain in the restricted scenarios is indeed possible under some conditions. We found that the strength of the dependence of two variables in the data-generating process as well as the regularization term of SNT have an influence on whether the attack succeeds or not. This is illustrated in Figure \ref{fig:success_vs_failure_full}, exact numbers of success rates can be obtained in Table \ref{tab:success_reates}. \\
Proceeding with attacks on forks, we obtain that converting a fork to a chain is possible with a linear number of attack trials. For example, assuming a fork $X_1 \leftarrow X_2 \rightarrow X_3$ as a data-generating process and data $\mathbf{X}$ from this process, we can scale $\mathbf{x}_3$ as often as we need, run the attack and obtain the graph produced by SNT. If the graph matches our expectation, we can be sure that we found a scale that forces SNT to predict a chain instead of a fork since SNT is fully deterministic given data $\mathbf{X}$ and fixed hyperparameters like the regularization. \\
Proceeding with attacking colliders, we consider the cases in which we don't have access to the collider-node's data itself. Thus, considering the 3-variable case, we only have to consider cases in which we have access to one of the outer nodes only. Now, a similar reasoning applies as in the fork-case: We only need a linear number of attacks in order to find a scale that allows us to flip one edge, effectively rendering a chain-structure in the resulting graph and an additional edge. For example, assume data $\mathbf{X}$ coming from a collider $X_1 \rightarrow X_2 \leftarrow X_3$. If we had access to $\mathbf{X}(\{1\})$ or $\mathbf{X}(\{3\})$ we only would need a linear number of trials to make our attack described in Section \ref{subsec:imperfect}.
\paragraph{Correlation between Model-MSE and Varsortability}
In order to further examine the properties of the relationship between varsortability and the MSE of a model found by SNT, we generated data according to a 3-node chain-graph $G = X_1 \rightarrow X_2 \rightarrow X_3$ as described above. Further we constructed a list of all possible 3-node DAGs. Since SNT is restricted to find DAGs (assuming the DAG-ness constrained equals $0$), the prediction of SNT is guaranteed to be included in this list. For each DAG we then fitted as set of linear models s.t. each linear model describes one variable in the DAG, e.g. for the chain graph $G$ from above we would fit a linear model describing $X_1 \rightarrow X_2$ and one linear model describing $X_2 \rightarrow X_3$. This is equivalent to what SNT does. We then can compute the Model-MSE of a graph $G$ with adjacency $\mathbf{W}$ (Model MSE) by:
\begin{equation}
    \begin{split}
          \MMSE(\mathbf{W}, \mathbf{X}) = & \sum_{i \in U} \Var(\mathbf{X}^T_i)  \\ & + \sum_{i \in N} \MSE(\mathbf{X}^T_i, \mathbf{X}, \mathbf{W})
    \end{split}
\end{equation}
Here, $\mathbf{X}^T_i$ refers to the data of variable $i$, $U$ is the set of exogenous variables in $G$ and $N$ is the set of endogenous variables in $G$. \\
The MMSE and varsortability are then computed for each DAG in the list of $25$ $3$-node DAGs and a corresponding set of linear models. We found that the MMSE and the varsortability-score indeed are correlated as shown in Figure \ref{fig:var_mse_correlation}, thus supporting the conjecture that the optimal Model-MSE of a graph $G$ is equivalent to varosrtability being $1$ w.r.t $G$. Additionally, one can see that more complex models tend to achieve a MMSE. This makes sense since complex models will capture more dependencies which is reflected in a lower MMSE. The last finding can be understood as an encouragement of SNT's usage of regularization in order to remove unnecessary (and possible false positive) dependencies found.

\section{Discussion and Conclusion}
We have confirmed that SNT is sensible for the scale of data used as input. Additionally we have shown that it is possible to perfectly control the output of SNT if we have full control of the data passed to SNT. This is possible by applying simple scaling operations on the data, thus there is not much computation power needed to employ such attacks against SNT. The empirical results were theoretically justified for perfect attack scenarios. \\
Our considerations of non-perfect attack-scenarios, i.e. the attacker does not have full access to the data, have empirically shown that it is still possible to successfully perform some kinds of attacks. However, since scale-information exploited by SNT cannot be fully manipulated, the success or failure of our attacks depend on the choice of hyperparameters chosen for SNT and the strength of dependence of the variables under attack.

\paragraph{Implications on use of SNT}
As already shown theoretically by other works, we confirmed that SNT is not a causal method since it derives the causal structure based on variance-properties of the input-data. Thus, SNT should not be used as a causal discovery method. Even the use of SNT as a structure learning method might be questioned since our attacks have shown that the resulting graph is fully controllable w.r.t. the dependencies in the data. 

\paragraph{Future Work}
One could extend the considerations from 3-node-systems to $n$-node-systems and see if our theoretical and empirical results still apply. In general, theoretical proofs about $n$-node cases are still lacking completely. An interesting future direction is applying our results on similar continuous optimization-approaches such as non-linear data-generating systems. Another open question is if there are loss-functions that can be used instead of least square based losses protecting SNT (and possibly other methods) from our attacks. Since in last years several methods were proposed to make neural networks more "causal" using similar approaches to SNT, one could examine if and to what extent these methods suffer from data-rescaling and our attacks.

\subsubsection*{Acknowledgments}
This work was supported by the ICT-48 Network of AI Research Excellence Center ``TAILOR'' (EU Horizon 2020, GA No 952215), the Nexplore Collaboration Lab ``AI in Construction'' (AICO) and by the Federal Ministry of Education and Research (BMBF; project ``PlexPlain'', FKZ 01IS19081). It benefited from the Hessian research priority programme LOEWE within the project WhiteBox and the HMWK cluster project ``The Third Wave of AI'' (3AI).
\bibliography{tant}

\clearpage

\newpage
\appendix

\section{Appendix}
We make use of this supplementary material to further extend on details regarding the content of the main paper ``Tearing Aparat NOTEARS.''

\subsection{Proof of Proposition 2}\label{proof:prop2}
\begin{proof}
First we note that due to Prop.\ref{prop:ReisachProof}, SNT will prefer selecting an edge $X_2 \leftarrow X_3$ instead of $X_2 \rightarrow X_3$. Now, assume to the contrary that $X_1 \rightarrow X_2 \leftarrow X_3$ is indeed the optimum i.e., the MSE is minimal. We know that the minimal MSE is proportional to the mutual information (MI) between $(X_1,X_3)$. The attack involves a scaling of $X_2$ such that the edge between $(X_2,X_3)$ flips, rendering the MI the same. However, the introduction of a collider renders $(X_1,X_3)$ suddenly independent given the Markov condition, effectively setting the MI to zero. Therefore, the pure collider cannot have been the optimal MSE.
\end{proof}

\subsection{Proof: Model-MSE as a sequence of Linear Model MSEs}\label{proof:modelmse_as_mse_and_vars}
In the following we will show that the Model-MSE (MMSE) minimized by SNT can be written in terms of regular linear regression MSEs and the variances of all exogenous variables. Note that we only allow for linear dependencies among the variables. \\
Given $n$ samples $\mathsf{X}$ where each sample comes from a zero-centered Gaussian distribution $\mathcal{N}$ over $d$ variables, SNT aims to minimize MMSE which is defined as:
\begin{equation*}
    \MMSE(\mathbf{W}, \mathbf{X}) = \frac{1}{2n} || \mathbf{X} - \mathbf{X}\mathbf{W} ||^2_F + h(\mathbf{W}) + \lambda ||\mathbf{W}||_1
\end{equation*}
Here, $\mathbf{W}$ is a $d \times d$-dimensional weight-matrix learned by SNT reflecting the dependency-structure of $\mathcal{N}$ as a DAG. In the following we will assume that $\lambda = 0$ and $\mathbf{W}$ is a DAG, that is we have no regularization and $h(\mathbf{W}) = 0$ holds. It then remains to show that $\frac{1}{2n} || \mathbf{X} - \mathbf{X}\mathbf{W} ||^2_F$ can be represented in terms of regular MSEs of a set of linear models and the variances of exogenous variables. Writing out the Frobenius-norm in this minimization problem we obtain:
\begin{align*}
    \MMSE(\mathbf{W}, \mathbf{X}) & = \frac{1}{2n} \sqrt{\sum_{i=1}^{n} \sum_{j=1}^d \big(\mathbf{X} - \mathbf{X} \mathbf{W}\big)^2_{i_j}} \\
    & = \frac{1}{2n} \sqrt{\sum_{j=1}^d ||\mathbf{X}^T_j - \mathbf{X} \mathbf{W}^T_j||^2} \\
    & \propto \frac{1}{2n} \sum_{j=1}^d ||\mathbf{X}^T_j - \mathbf{X} \mathbf{W}^T_j||^2
\end{align*}
Thus we can express the MMSE as the sum of $d$ independent MSE-terms:
\begin{align*}
    \MSE(\mathbf{X}^T_j, \mathbf{W}, \mathbf{X}) & = \frac{1}{2n} ||\mathbf{X}^T_j - \mathbf{X} \mathbf{W}^T_j||^2 \\
    & = \sum_{i=1}^n (\mathbf{X}_{i_j} - \mathbf{X}_i \mathbf{W}^T_j)^2
\end{align*}
There are two cases we have to consider: (1) The weight-vector $\mathbf{W}^T_j = \mathbf{0}$, i.e. a node $X_j$ in the graph represented by $\mathbf{W}$ has no parents, and (2) $\mathbf{W}^T_j \neq \mathbf{0}$, that is, a node $X_j$ in has parents in the graph.
In case of (1) we can shorten the MSE to:
\begin{align*}
   \MSE(\mathbf{X}^T_j, \mathbf{W}, \mathbf{X}) & = \sum_{i=1}^n (\mathbf{X}_{i_j} - \mathbf{X}_i \mathbf{W}^T_j)^2 \\
   & = \sum_{i=1}^n \mathbf{X}_{i_j}^2
    = \Var(\mathbf{X}^T_j)
\end{align*}
Thus, the MMSE can be expressed as follows:
\begin{align*}
    \MMSE(\mathbf{W}, \mathbf{X}) & = \sum_{i=1}^d \mathrm{I}_{\mathbf{W}^T_i = 0} \Var(\mathbf{X}^T_i) \\ 
    & \quad + (1 - \mathrm{I}_{\mathbf{W}^T_i = 0}) \MSE(\mathbf{X}^T_i, \mathbf{X}, \mathbf{W}) \\
    & = \sum_{i \in Z} \Var(\mathbf{X}^T_i) + \sum_{i \in N} \MSE(\mathbf{X}^T_i, \mathbf{X}, \mathbf{W})
\end{align*}
Here, $\mathrm{I}_{\mathbf{W}^T_j = 0}$ is the indicator function which equals $1$ iff $\mathbf{W}^T_j = 0$, $Z$ is the set of variable indices for which $\mathbf{W}^T_j = 0$ holds and $N = \{1, \dots, d\} \setminus Z$. \qed

\subsection{Further Details}\label{sec:figures}
\begin{figure}[ht!]
     \centering
     \includegraphics[width=.5\textwidth]{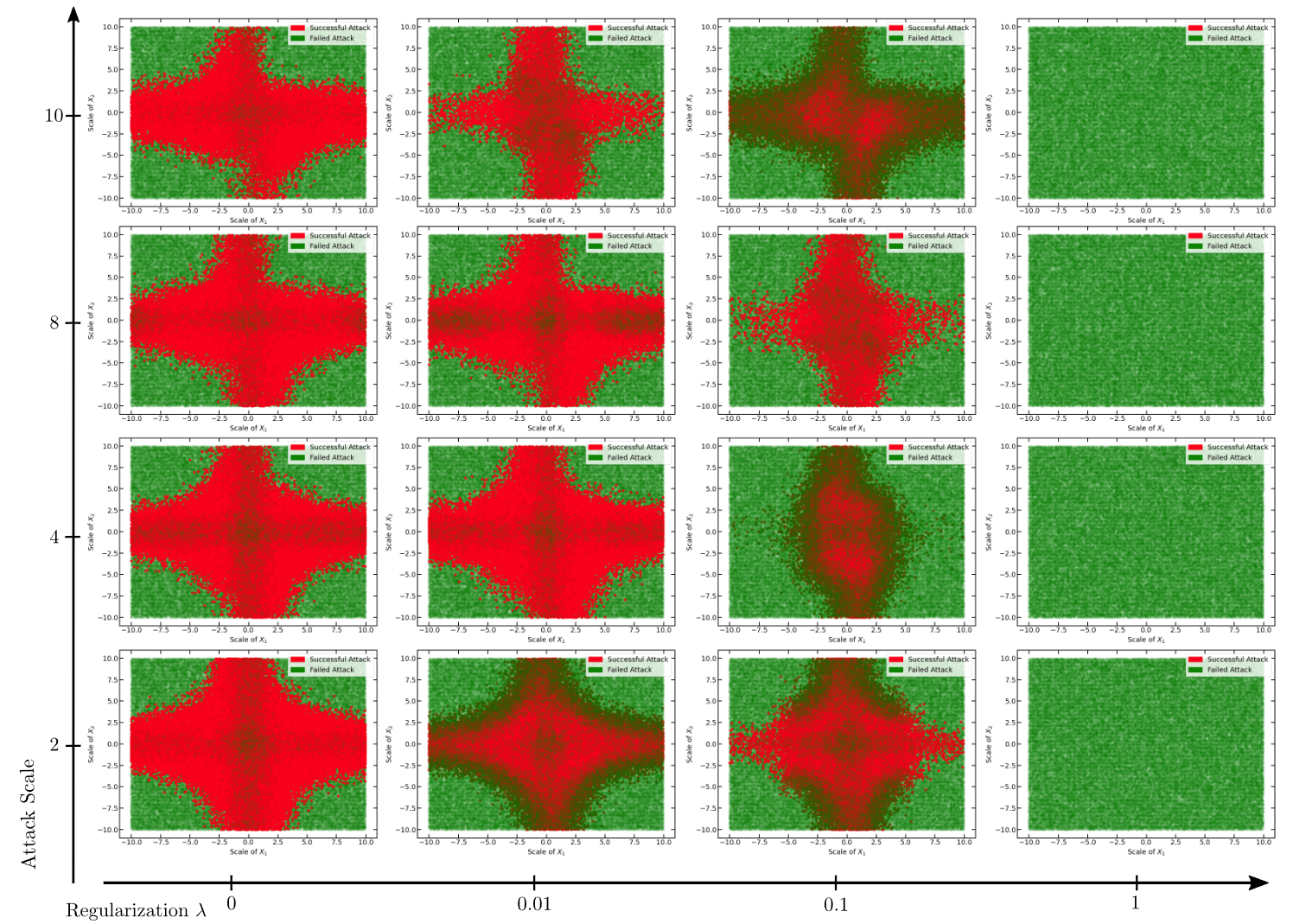}
     \caption{\textbf{Success Ratios of chain-reversals in imperfect scenarios.} The success ratios of reversing a chain by our attack in the imperfect scenario shows a significant dependency between the choice of thescaling factor used in the attack, the regularization term $\lambda$ used for SNT and the likelihood to succeed with the attack.}
     \label{fig:success_vs_failure_full}
     \vspace{-.5cm}
 \end{figure}

\textbf{Scale and Regularization.} In Fig.\ref{fig:success_vs_failure_full} we present a brief ablation on attack scale and regularization.
 
\textbf{Attacking Chains.} First we will show how SNT can be attacked if data $\mathcal{D}$ comes from a causal graph with a chain-structure. There are three ways we can attack a chain: (1) Reverting the chain, (2) converting a chain into a fork and (3) introducing a collider in the chain. The following paragraphs will describe the attacks in detail.

\textit{Reverting Chains.} Forcing SNT to revert a chain can be done by making the variables in $\mathcal{D}$ fully varsortable w.r.t.\ the graph we want SNT to produce. For example, assume that data comes from a chain $X_1 \rightarrow X_2 \rightarrow X_3$. If we scale the data s.t. $\Var(X_1) > \Var(X_2) > \Var(X_3)$ and apply SNT on the attacked dataset, it will predict a chain $X_1 \leftarrow X_2 \leftarrow X_3$.

\textit{Converting Chains to Forks.} Converting a chain to a fork can be done in a similar way as reverting a chain. Again, assume data comes from a chain $X_1 \rightarrow X_2 \rightarrow X_3$. In order to force SNT to predict the fork $X_1 \leftarrow X_2 \rightarrow X_3$, we have to scale the data s.t. $\Var(X_1) > \Var(X_2) < \Var(X_3)$.

\textit{Converting Chains to Colliders.} Creating a collider out of a chain works similar again. Assuming the same data-generating process as above, we just have to scale data s.t. $\Var(X_2) < \Var(X_1)$ and $\Var(X_3) < \Var(X_2)$ to obtain a collider in $X_2$. However, SNT will introduce an additional edge between $X_1$ and $X_3$. This is because in the data-generating process these two variables are dependent on each other over $X_2$. If we turn the edges s.t. SNT builds a collider at $X_2$, SNT would not respect this dependence. Thus, in order to still minimize MSE, SNT is forced to add this additional edge to account for his dependence.

\textbf{Attacking Forks.} The second structure we attack are forks. Here we can perform two attacks: (1) Converting a fork to a chain and (2) converting a fork into a collider. There is no third attack possible since this would require us to change the dependencies in the data. E.g. given a fork $X_1 \leftarrow X_2 \rightarrow X_3$, we cannot change the fork's origin node to $X_2 \leftarrow X_1 \rightarrow X_3$ since this would require us to replace the conditional independence $X_3 \indep X_1 | X_2$ by $X_2 \indep X_3 | X_1$. 

\textit{Converting Forks to Chains.} Converting a fork into a chain works again by making the data varsortable w.r.t. the chain we wish SNT to predict. That is, given a fork $X_1 \leftarrow X_2 \rightarrow X_3$, we have to scale the data s.t. $\Var(X_1) < \Var(X_2) < \Var(X_3)$ in order to force SNT to predict $X_1 \rightarrow X_2 \rightarrow X_3$. 

\textit{Converting Forks to Colliders.} Forcing SNT to predict a collider if data comes from a fork works exactly as in the case of converting a chain into a collider. Given data from a fork $X_1 \leftarrow X_2 \rightarrow X_3$, we have to scale the data s.t. $\Var(X_2) > \Var(X_1)$ and $\Var(X_2) > \Var(X_3)$ in order to make SNT predicting a collider. Since this fork shares the same independence-statements as a chain $X_1 \rightarrow X_2 \rightarrow X_3$, SNT will again place an additional edge between $X_1$ and $X_3$ since we would remove this with a collider. The proof is the same as in Proposition \ref{prop:Chain2Collider}.

\textbf{Attacking Colliders.} Colliders can be attacked in two ways: (1) Convert a collider into a chain and (2) convert a collider into a fork. Another attack is not possible since we then would have to change the (in)dependence-statements found in the data again.

\textit{Converting Colliders to Chains.} Assuming a collider $X_1 \rightarrow X_2 \leftarrow X_3$, we can employ an attack on the data s.t. SNT will predict the graph $X_1 \rightarrow X_2 \rightarrow X_3$ with an additional edge between $X_1$ and $X_3$ by scaling data s.t. $\Var(X_1) < \Var(X_2) < \Var(X_3$. The additional edge is added in order to account for the conditional dependence between $X_1$ and $X_3$ if we observe $X_2$. 

\textit{Converting Colliders to Forks.} Converting a collider $X_1 \rightarrow X_2 \leftarrow X_3$ into a fork $X_1 \leftarrow X_2 \rightarrow X_3$ can be achieved by scaling data s.t. $\Var(X_2) < \Var(X_1)$ and $\Var(X_1) < \Var(X_3)$. Additionally, SNT will again add an additional edge between $X_1$ and $X_3$ to account for the conditional dependence as before for collider to chain.

\textbf{Success Ratios.} In Tab.\ref{tab:success_reates} we present a brief ablation on attack success ratios.
\begin{table}[!h]
\centering
\caption{\textbf{Success Ratios of chain-reversals in imperfect scenarios.} The success ratios of reversing a chain by our attack in the imperfect scenario shows a significant dependency between the choice of thescaling factor used in the attack, the regularization term $\lambda$ used for SNT and the likelihood to succeed with the attack.}
\label{tab:success_reates}
\begin{tabular}{l||l|llll} 
\toprule
                              & \multicolumn{1}{l}{} & \multicolumn{4}{l}{Regularization $\lambda$}  \\ 
\hhline{=::=====}
                              &                      & $0$    & $0.01$ & $0.1$  & $1$                \\ 
\cline{2-6}
\multirow{4}{*}{\rotatebox[origin=c]{90}{\parbox[c]{1cm}{\centering Attack Scale}}} & $2$                  &$0.35$ & $0.32$ & $0.23$ & $0.$               \\
                              & $4$                  & $0.30$ & $0.28$ & $0.15$ & $0.$               \\
                              & $8$                  & $0.23$ & $0.21$ & $0.10$ & $0.$               \\
                              & $10$                 & $0.19$ & $0.19$ & $0.10$ & $0.$               \\
\bottomrule
\end{tabular}
\end{table}

\subsection{Technical Details}\label{sec:tech_details}
Our code is available at \url{https://anonymous.4open.science/r/TANT-8B50/}. In our experiments we considered 3-node-SCMs only. In each attack we sampled $10000$ samples from a Gaussian distribution for each noise term. Each endogenous node was computed by a linear combination of its parents and an additive Gaussian noise term. Our attacks used the original NOTEARS implementation. For each experiments in the imperfect attack-setting we sampled each noise term only once in order to perform the attack on the same data for different attack-scales and different values of $\lambda$. The data is available in our repository for reproducibility. All attacks were performed on a regular laptop machine with a AMD Ryzen 7 CPU and 16GB RAM. 

\end{document}